\documentclass[10pt,twocolumn,letterpaper]{article}

\usepackage{cvpr}              

\definecolor{cvprblue}{rgb}{0.21,0.49,0.74}
\usepackage[pagebackref,breaklinks,colorlinks,allcolors=cvprblue]{hyperref}

\def\paperID{38727} 
\def\confName{CVPR}
\def\confYear{2026}

\title{EggHand: A Multimodal Foundation Model\\for Egocentric Hand Pose Forecasting}

\author{
Jaeyoung Choi\thanks{Equal contribution.},\;\;Hyeondong Kim\footnotemark[1],\;\;Yujin Kim,\;Daehee Park\thanks{Corresponding author.}\\
DGIST, Republic of Korea\\
\texttt{\{cjyoung, hdkim, yujin2110, dhpark\}@dgist.ac.kr}
}
\begin{document}


\maketitle

\begin{abstract}
Forecasting future 3D hand pose sequences from egocentric video is essential for understanding human intention and enabling embodied applications such as AR/VR assistance and human–robot interaction.
However, this task remains a highly challenging problem because egocentric hand motion is driven by complex human intent, exhibits highly dexterous articulations, and is observed under drastic viewpoint shifts induced by ego-motion.
In this work, we introduce \textbf{EggHand}, a foundation-model-based framework for egocentric hand pose forecasting that unifies multimodal semantic reasoning with dynamic motion modeling.
Our approach couples an action decoder from a Vision-Language-Action (VLA) model, which captures the structured temporal dynamics of hand motion, with an egocentric video–text encoder that provides viewpoint-aware contextual information learned from large-scale first-person video.
Together, these components overcome the brittleness of generic visual encoders under ego-motion and enable joint reasoning over motion, context, and high-level intent—without relying on body pose or external tracking.
Experiments on the EgoExo4D dataset show that EggHand sets a new state of the art in forecasting accuracy, remains robust under severe ego-motion, and further enables controllable prediction via language-based task prompts. Project page: \href{https://jyoun9.github.io/EggHand}{https://jyoun9.github.io/EggHand}

\end{abstract}

\section{Introduction}
\label{sec:intro}
\begin{figure}[t]
  \centering
   \includegraphics[width=1.0\linewidth]{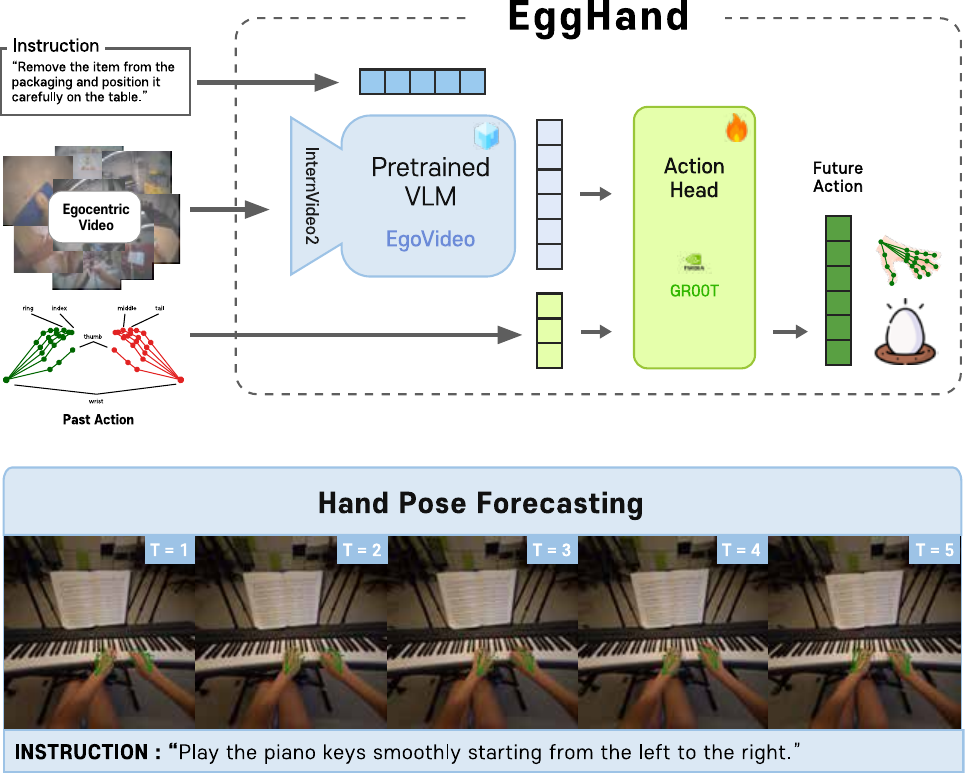}

   \caption{Overview of EggHand, the proposed framework for language-conditioned 3D hand pose forecasting. The model forecasts future 3D hand poses conditioned on past 3D hand positions, the input video, and text instructions. The hand pose forecasting visualization overlays the forecasted poses onto the 2D video frames for qualitative evaluation.}
   \label{fig:onecol}
\end{figure}

Egocentric hand pose forecasting aims to predict how articulated 3D hand poses will evolve over time from first-person observations. 
This capability is essential for understanding human intention, anticipating future interactions, and enabling embodied applications such as AR/VR assistance and human–robot collaboration~\cite{chang2023look,grauman2022ego4d}.  
Predicting how hands will move in 3D from egocentric video is thus a critical prerequisite for proactive planning in embodied AI systems~\cite{hatano2025invisible,qi2023diverse}.
The problem remains challenging because egocentric hand pose reflects complex human intentions, highly dexterous movements, and continuously changing viewpoints caused by ego motion~\cite{liu2022joint,hatano2024emag,hatano2025invisible}.

Recent works have explored egocentric hand forecasting by modeling visual dynamics, hand–object interactions, or full-body cues~\cite{ma2025novel, zhou2024megohand}. 
EMAG~\cite{hatano2024emag} introduced ego-motion-aware learning to stabilize 2D trajectory prediction under camera motion, while EgoH4~\cite{hatano2025invisible} leveraged body-pose estimation to infer future 3D hand poses even when hands are not visible. 
Other transformer-based approaches capture temporal continuity and contact reasoning to predict future gestures or interactions~\cite{ji2024egocentric,bao2024handsonvlm,he2025gaze, ye2025predicting}. 
Although these methods have improved short-term prediction accuracy, they remain limited by task-specific training and poor generalization across diverse environments. 
In particular, most models rely on low-level motion cues without explicitly modeling the high-level semantics or human intentions that drive hand pose evolution.
Several approaches assume access to unrealistic inputs such as full-body pose, 3D tracking, or manually provided interaction cues.

To address these limitations, we utilize foundation models with strong multimodal reasoning capabilities.
Motivated by this, we introduce EggHand(Fig.~\ref{fig:onecol}), a multimodal foundation model for egocentric hand pose forecasting.

While several motion forecasting domains, such as trajectory prediction~\cite{gao2024multi, gao2025omnitraj}, have recently adopted foundation models for generalizable spatiotemporal reasoning, such approaches have not to date been explored in egocentric 3D hand pose forecasting. 
Recent works commonly use vision–language models (VLMs) as decoders for 2D trajectory prediction~\cite{bae2024can,hwang2024emma,zhang2024instruct}. 
However, unlike low-dimensional 2D trajectories, hand pose involves high-dimensional articulated dynamics and strict kinematic constraints, making it challenging to directly reuse a generic VLM decoder for accurate hand pose forecasting~\cite{bao2024handsonvlm, WANG2026104508}. 
Therefore, instead of a generic VLM decoder, we reuse the action decoder from a vision–language–action (VLA) model, which is pretrained to generate physically meaningful action trajectories and to capture how motions evolve over time~\cite{bjorck2025gr00t, brohan2022rt, brohan2023rt}. 
However, existing VLAs are not specialized for egocentric scenarios, where ego motion continuously changes the viewpoint and the spatial relationships between the hands and objects. 
To handle this, we incorporate an egocentric video–text foundation encoder, which provides viewpoint-aware contextual grounding learned from large-scale first-person video datasets~\cite{grauman2022ego4d,grauman2024ego,ma2024nymeria}. 
This design allows EggHand to combine egocentric visual context with action-structured decoding, rather than relying purely on low-level motion cues.

Finally, we propose a geometry-aware training objective that integrates multimodal embeddings with physically grounded 3D hand supervision.
Our loss function jointly enforces (i) wrist-relative spatial alignment, (ii) inter-joint geometric consistency, and (iii) temporally coherent future prediction.
This formulation directly targets common failure modes of egocentric hand pose forecasting, such as global drift under ego motion, anatomically implausible poses, and temporally jittery trajectories, enabling stable and realistic forecasts under diverse interaction contexts.

Our method is validated on real-world EgoExo4D dataset~\cite{grauman2024ego}, achieving superior forecasting and generalization performance compared to previous approaches.
It remains robust under large ego motion and further enables controllable hand pose generation through task-level prompt conditioning.
This opens possibilities for broader embodied applications in AR/VR and robotics.

Our contributions are summarized as follows:
\begin{itemize}
    \item We propose a new foundation-model that couples a VLA action decoder with an egocentric video–text encoder, yielding a unified model that is robust to ego motion while jointly understanding high-level scene semantics and the complex dynamics of 3D hand poses.
    \item We introduce a geometry-aware training objective that jointly optimizes absolute pose accuracy, relative wrist-anchored stability, and intra-hand geometric consistency, leading to stable and anatomically plausible hand pose forecasts in egocentric settings.
    \item We show that our approach achieves state-of-the-art performance on real-world egocentric datasets (EgoExo4D), and further supports controllable hand pose prediction driven by high-level task prompts.
\end{itemize}

\section{Related Works}

\subsection{Human Motion Understanding} 
Human motion understanding is a fundamental problem in computer vision and robotics, aiming to model how humans perceive, act, and interact in the physical world~\cite{aggarwal2011human,poppe2010survey,zhu2024human}. 
Research in this area has progressed along two complementary directions: semantic action understanding and geometric motion modeling. 
Semantic approaches focus on predicting what a person intends to do, such as recognizing or anticipating future actions from visual and contextual cues~\cite{simonyan2014two,feichtenhofer2019slowfast, wang2023memory,mittal2024can}. 
Geometric approaches, on the other hand, model how the body or hands move through space, estimating articulated poses, 2D trajectories, or 3D motion patterns~\cite{cao2017realtime,sun2019deep,martinez2017human,pavllo2019quaternet,mao2022weakly,shi2022motion}. 
Both perspectives capture different aspects of human behavior—semantics explain why a motion occurs, while geometry describes how it unfolds. 
For comprehensive human motion forecasting, it is essential to integrate these two perspectives, reasoning jointly about the intent that drives an action and the physical dynamics that realize it over time~\cite{li2021action,gong2022future,fernando2021anticipating}.
Recent works further highlight the importance of multi-modal cues such as language.
However, most of these methods operate in third-person settings, often with full-body skeletons or coarse trajectories. 
In this work, we instantiate this perspective at the level of egocentric hand poses, where subtle finger motions and strong viewpoint changes make the coupling between intent and kinematics particularly challenging.

\subsection{Egocentric Hand Motion Forecasting}
Building upon these ideas, recent research has begun to explore egocentric hand motion forecasting, which predicts how human hands will move in first-person video~\cite{damen2022rescaling,kwon2021h2o,ohkawa2023assemblyhands}.
EMAG~\cite{hatano2024emag} addresses ego-motion compensation via homography-based representations and leverages multiple modalities such as optical flow and object trajectories to improve cross-scene generalization. EgoH4~\cite{hatano2025invisible} instead exploits ego-body pose estimation to forecast 3D hand positions, even when hands are out of view during observation.
While these approaches capture short-term dynamics and spatial consistency, they remain limited in reasoning about human intention or semantic context. 
Many rely on low-level motion features or handcrafted priors, which can lead to limited generalization under new activities or environments. 
In egocentric settings, frequent viewpoint changes and complex interactions further amplify these challenges~\cite{qi2023diverse}. 
To overcome these limitations, hand motion forecasting should be approached as a contextual reasoning problem that connects low-level motion with high-level human goals and environmental context.

\subsection{Foundation Models for Motion and Embodied Forecasting}
Recent advances in motion understanding have introduced the use of foundation models—particularly vision-language models (VLMs)—to enhance generalization and semantic reasoning~\cite{radford2021learning,brohan2023rt}. 

Approaches such as MotionGPT~\cite{zhang2024motiongpt} and GR00T~\cite{bjorck2025gr00t} demonstrate that foundation models can generate semantically grounded motion outputs. In the trajectory prediction domain, OmniTraj~\cite{gao2025omnitraj} shows that pre-training on heterogeneous trajectory data with explicit temporal conditioning enables zero-shot transfer across datasets with varying frame rates.

However, these models primarily focus on full-body or agent-level motion and overlook the fine-grained dexterity and viewpoint variations that characterize hand-level egocentric settings. 
Moreover, most rely solely on vision-language alignment and do not explicitly encode the dynamic structure of actions. 
A parallel line of work has explored using VLMs and LLMs as decoders for 2D trajectory prediction and embodied decision-making, where language prompts specify goals and the model outputs coarse future paths. 
While these systems benefit from strong semantic priors, their decoders are not designed to respect articulated hand kinematics, nor to remain stable under severe ego motion and occlusion. 
To address these gaps, our work combines the action decoder from GR00T~\cite{bjorck2025gr00t}, which captures complex motion dynamics, with the egocentric video-text encoder from EgoVideo~\cite{pei2024egovideo}, which provides viewpoint‑aware contextual grounding learned from large-scale first-person data. 
\section{Method}

\subsection{Problem definition}
\label{sec:problem_definition}
We study egocentric hand motion forecasting from first-person video.
Let $T_{\mathrm{obs}}$ be the observation length and $T_{\mathrm{fut}}$ the prediction horizon.
Time indices are $t\in\{-T_{\mathrm{obs}},\ldots,0\}$ for observed frames and $t\in\{1,\ldots,T_{\mathrm{fut}}\}$ for future frames.

Let $\mathcal{J}$ be the set of hand keypoints with $|\mathcal{J}|=J$.
The multimodal input is
\begin{equation}
\mathcal{X}
=
\big\{
I_{-T_{\mathrm{obs}}:0},\;
\mathbf{P}_{-T_{\mathrm{obs}}:0},\;
c
\big\},
\end{equation}

where $I_t$ is RGB frame at $t$, 
$\mathbf{P}_{-T_{\mathrm{obs}}:0}\!\in\!\mathbb{R}^{(T_{\mathrm{obs}}+1)\times J\times 3}$ is the past hand motion with
$\mathbf{P}_{t}=(\mathbf{p}_{1,t},\ldots,\mathbf{p}_{J,t})$ and $\mathbf{p}_{j,t}\in\mathbb{R}^{3}$,
and $c$ is a text description of the situation.

All 3D hand poses are represented in a normalized egocentric canonical frame, following EgoH4~\cite{hatano2025invisible}.

We express all joints relative to the first observation frame by transforming world-coordinate poses using the corresponding camera pose.
This canonicalization removes global motion while preserving the intrinsic structure of hand articulation.

Our model directly predicts future poses in this canonical egocentric space, without reverting to the global frame.

The output is the future hand motion tensor
\begin{equation}
Y
=
\mathbf{P}_{1:T_{\mathrm{fut}}}
\in
\mathbb{R}^{T_{\mathrm{fut}}\times J\times 3},
\quad
\mathbf{P}_{t}
=
(\mathbf{p}_{1,t},\ldots,\mathbf{p}_{J,t}).
\end{equation}

We learn the parameters $\theta$ of a conditional model $f_{\theta}$ that, given the environment and context encoded in $\mathcal{X}$, predicts future hand motion $Y$ faithful to the ground-truth trajectories and semantically consistent with $c$.
In practice, training uses supervised pairs $(\mathcal{X},\hat{Y})$ with point-wise 3D joint errors (e.g., MPJPE) as the primary signal.
Crucially, egocentric image streams exhibit severe ego-motion and frequent viewpoint shifts that make generic vision encoders brittle;
therefore we employ an egocentric video–text encoder to obtain viewpoint-aware context and a VLA-style action decoder to preserve action structure during forecasting.

\begin{figure}[t]
\centering
    \includegraphics[width=1.0\linewidth]{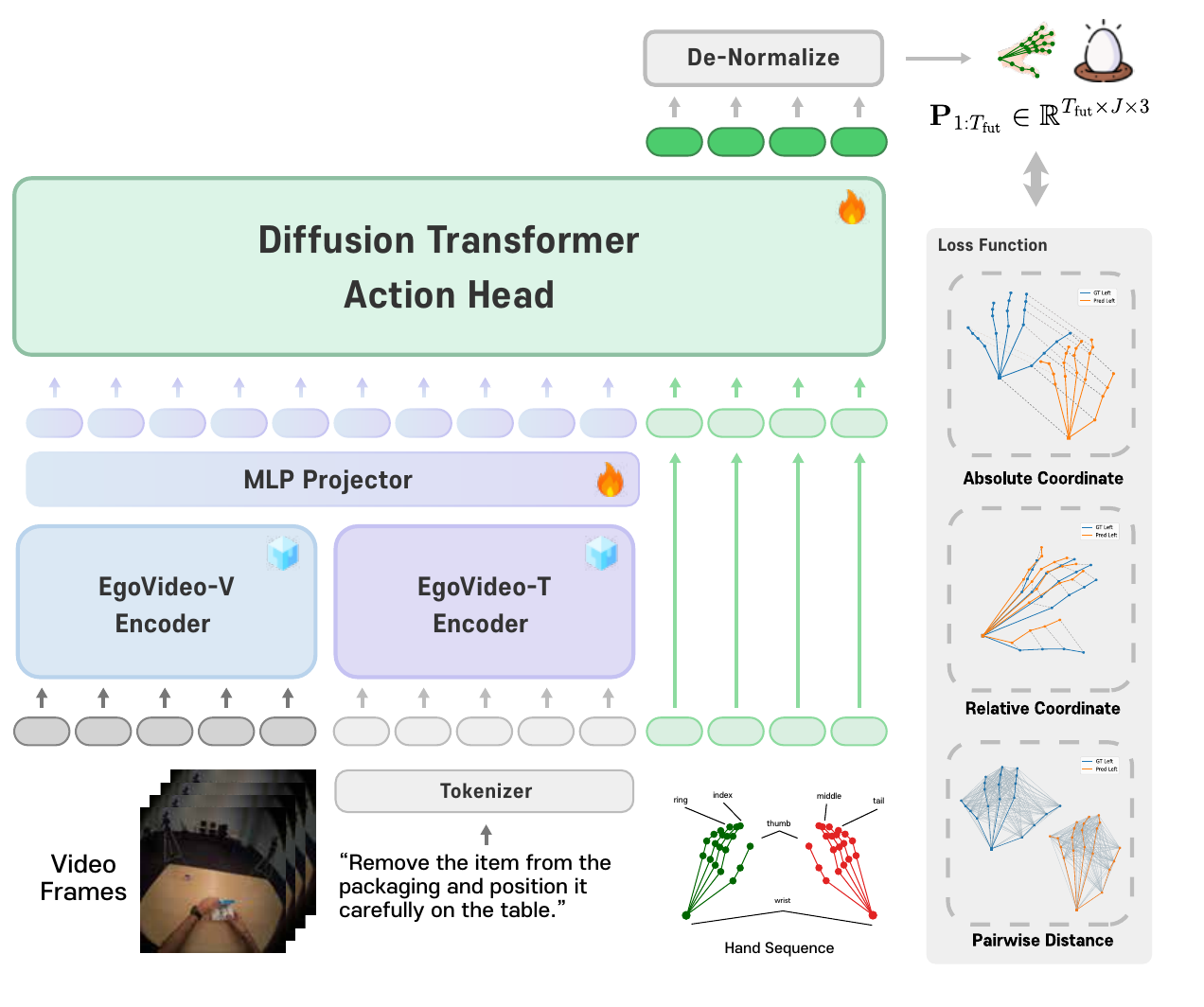}
   \caption{Framework of the proposed VLA architecture. The EgoVideo VLM extracts embeddings from both video and text, and the GR00T action decoder predicts the hand pose. During training, the model is  optimized with our proposed loss functions.}
   \label{fig:method_main}
\end{figure}
%
%

\subsection{Overview}
\label{sec:method_overview}
Given the problem in Sec.~\ref{sec:problem_definition}, our approach transfers complementary knowledge from two foundational models with different pretraining goals:
an egocentric video–text encoder (\textbf{EgoVideo})~\cite{pei2024egovideo} for viewpoint-aware context
and a vision–language–action decoder (\textbf{GR00T})~\cite{bjorck2025gr00t} for action-structured generation.
The method has two pillars: a cross-model fusion that forms a clean interface between the encoders and the action decoder (Sec.~\ref{sec:cross_model_fusion}), and a training objective that aligns semantics and future 3D hand motion while preserving egocentric stability (Sec.~\ref{sec:training_objectives}).

\subsection{Cross-model Fusion}
\label{sec:cross_model_fusion}

Our rationale is two-fold as shown in Fig.~\ref{fig:method_main}:
(i) egocentric video streams exhibit severe ego-motion and fast viewpoint changes, for which a generic vision encoder is brittle. To address this,
an egocentric video–text encoder (EgoVideo) supplies viewpoint-aware context that is stable under such shifts.
(ii) Future hand motion is action-structured and semantically driven. Therefore,
a decoder of GR00T preserves the temporal structure of actions and enables text-conditioned control.
By coupling these two components through a cross-model fusion layer,
we obtain a viewpoint-aware yet action-coherent prediction framework for egocentric hand forecasting.

We keep the EgoVideo encoder frozen, and fine-tune the GR00T-based action decoder mainly through lightweight adapters and cross-attention layers.
This preserves the generalization ability of both pretrained components while allowing the fusion layers to specialize to egocentric 3D hand pose forecasting.

\subsubsection{Egocentric Video-Text Encoder}
\label{sec:egovideo_encoder}

Egocentric videos often involve unstable or rapidly moving cameras and contain detailed representations of complex hand–object interactions.
While recent VLMs demonstrate strong generalization across many video tasks, they are not specifically optimized for egocentric footage.  
To address this, we adopt the EgoVideo model, a foundation VLM pre-trained on large-scale egocentric datasets.  
EgoVideo offers viewpoint-aware visual representations and has achieved strong performance across multiple egocentric video benchmarks, making it well-suited for our setting.

Given an input egocentric video 
\begin{equation}
I_{-T_{\mathrm{obs}}:0} \in \mathbb{R}^{B\times C\times T_v\times H\times W},
\end{equation}

the EgoVideo visual encoder produces latent visual representations
\begin{equation}
V = f_v(I_{-T_{\mathrm{obs}}:0}) \in \mathbb{R}^{B\times L_{\text{raw}}\times D_v}.
\end{equation}

The text description $c$ is processed by the EgoVideo text encoder:
\begin{equation}
T = f_t(c) 
\in \mathbb{R}^{B\times L_t\times D_t}.
\end{equation}
The text can be obtained by sampling representative frames from $I_{-T_{\mathrm{obs}}:0}$ and generating a multimodal instruction that describes the egocentric context using a foundation model (e.g., GPT-4o-mini).
This allows the VLA to receive intent-aligned semantic cues directly from the video.

Both streams are projected into a shared latent space via lightweight adapters:
\begin{equation}
\tilde{V} = V W_v + b_v, 
\quad
\tilde{T} = T W_t + b_t,
\end{equation}
where $W_v \in \mathbb{R}^{D_v\times D}$ and $W_t \in \mathbb{R}^{D_t\times D}$.
Temporal and positional encodings are added:
\begin{equation}
\hat{V} = \tilde{V} + E_{L_v}^{(\mathrm{time})},
\qquad
\hat{T} = \tilde{T} + E_{L_t}^{(\mathrm{pos})}.
\end{equation}
The concatenated multimodal context is then formed as
\begin{equation}
C = \mathrm{Concat}(\hat{V}, \hat{T})
\in \mathbb{R}^{B\times (L_v + L_t)\times D}.
\end{equation}

This fused context $C$ serves as the input interface to the GR00T-based VLA module described below.


\subsubsection{Vision-Language-Action Decoder}
\label{sec:vla_decoder}

The GR00T-based VLA decoder receives the multimodal context $C$
and the observed hand states as inputs.
The observed 3D hand motion sequence
$\mathbf{P}_{-T_{\mathrm{obs}}:0}$,
already transformed into the normalized egocentric frame anchored at the first observation timestep,
is encoded via
\begin{equation}
S = f_s(\mathbf{P}_{-T_{\mathrm{obs}}:0})W_s + b_s
\in \mathbb{R}^{B\times (T_{\mathrm{obs}}+1)\times D_s}.
\end{equation}
A cross-attention encoder integrates both streams:
\begin{equation}
Z = \mathrm{CrossAttentionEncoder}(S, C),
\end{equation}
yielding a viewpoint-aware latent embedding that encodes temporal motion and semantic intent jointly.

The action decoder is initialized from the GR00T foundation model,
whose flow-matching pretraining endows the module with strong temporal action priors.
We fine-tune this decoder with supervised 3D joint losses using lightweight adapters,
resulting in deterministic and stable future hand motion predictions:
\begin{equation}
\hat{\mathbf{P}}_{1:T_{\mathrm{fut}}} = g_{\theta}(Z).
\end{equation}



All hand trajectories—both inputs $\mathbf{P}_{-T_{\mathrm{obs}}:0}$ and outputs $\hat{\mathbf{P}}_{1:T_{\mathrm{fut}}}$—are represented in the normalized egocentric frame at each time step during training and inference.
For visualization and downstream evaluation, all predicted poses are kept in the same egocentric motion frame used during training step.
This per-frame normalization strategy removes camera-pose variability and keeps the model’s prediction space consistently aligned with the egocentric view.


%
%
\subsection{Training Objectives}
\label{sec:training_objectives}

%
%

While the action head inherits flow-matching priors from GR00T’s pretraining, our fine-tuning is fully supervised, using 3D joint losses to achieve stable and physically meaningful motion prediction.

We train the model to predict future 3D hand joint trajectories by minimizing a composite loss that enforces both absolute and relative accuracy while preserving intra-hand geometric consistency.
Let $T = T_{\mathrm{fut}}$ denote the length of the forecast horizon for notational simplicity.
The overall training objective is formulated as:
\begin{equation}
\mathcal{L}_{\text{total}} 
= 
\lambda_{\text{abs}} \, \mathcal{L}_{\text{abs}} 
+ 
\lambda_{\text{rel}} \, \mathcal{L}_{\text{rel}} 
+ 
\lambda_{\text{pair}} \, \mathcal{L}_{\text{pair}},
\end{equation}
where $\lambda_{\text{abs}}$, $\lambda_{\text{rel}}$, and $\lambda_{\text{pair}}$ denote the weighting coefficients of each loss component.

\paragraph{Absolute Coordinate Loss.}
The absolute loss measures the point-wise deviation between predicted and ground-truth 3D joint positions in the egocentric coordinate frame:
\begin{equation}
\mathcal{L}_{\text{abs}}
=
\frac{1}{TJ}
\sum_{t=1}^{T}
\sum_{j=1}^{J}
\big\|
\hat{\mathbf{p}}_{j,t} - \mathbf{p}_{j,t}
\big\|_1.
\end{equation}
This term enforces accurate spatial alignment of all joints within the camera frame, directly optimizing the per-joint position error (MPJPE).

\paragraph{Relative Coordinate Loss.}
To enhance local coherence and mitigate global pose drift, we additionally impose a wrist-centered relative loss~\cite{matsune2024geometry}:
\begin{equation}
\mathcal{L}_{\text{rel}}
=
\frac{1}{TJ}
\sum_{t=1}^{T}
\sum_{j\in \mathcal{J}_{\text{hand}}}
\Big\|
\big(\hat{\mathbf{p}}_{j,t} - \hat{\mathbf{p}}_{\text{wrist},t}\big)
-
\big(\mathbf{p}_{j,t} - \mathbf{p}_{\text{wrist},t}\big)
\Big\|_1.
\end{equation}
This formulation preserves the internal configuration of hand joints relative to their wrist anchor, which improves robustness under egocentric viewpoint changes.

\paragraph{Pairwise Distance Loss.}
Finally, to preserve intra-hand geometric integrity, we adopt a pairwise distance regularizer~\cite{spurr2020weakly}:
\begin{equation}
\mathcal{L}_{\text{pair}}
=
\frac{1}{T|\mathcal{S}|}
\sum_{t=1}^{T}
\sum_{(i,j)\in \mathcal{S}}
\big(
\|
\hat{\mathbf{p}}_{i,t} - \hat{\mathbf{p}}_{j,t}
\|_2
-
\|
\mathbf{p}_{i,t} - \mathbf{p}_{j,t}
\|_2
\big)^2,
\end{equation}
where $\mathcal{S}$ represents the set of valid intra-hand joint pairs.
Unlike coordinate-based losses, this distance-based formulation regularizes the model in a relational space, encouraging geometric consistency across joint pairs.

Following previous findings that $L_1$ loss performs better for human joint prediction~\cite{Sun2017comp}, we apply $L_1$ loss for coordinate-level supervision and $L_2$ loss for pairwise geometric regularization to fine-adjust for local geometric inconsistencies between joints.

\section{Experiments}

\label{sec:experiments}

This section describes our experimental setup, including datasets, preprocessing, baselines, implementation details, and evaluation metrics.

\subsection{Experiment Setup}
\label{sec:dataset_setup}

\paragraph{Dataset.}
We evaluate our method using the EgoExo4D dataset~\cite{grauman2024ego}, which contains synchronized egocentric RGB video streams and calibrated 3D hand pose annotations collected from Meta Aria devices. 
Each clip contains RGB frames at 10 fps, 3D hand joint annotations for 42 joints (21 joints per hand, including the wrists), and per-joint availability flags derived from the dataset’s official annotations. 
All distances are measured in meters. 
We do not use any exocentric views, whole-body tracking, or other external 3D signals.

\paragraph{Motion Coordinate Frame.}
All 3D hand poses are represented in a normalized egocentric canonical frame.
Following EgoH4~\cite{hatano2025invisible}, we define a fixed reference coordinate system at the first observation timestep $-T_{\text{obs}}$.

Specifically, raw world-coordinate joints are transformed using the camera pose at $-T_{\text{obs}}$, removing global translation and rotation while preserving gravity direction.
All subsequent poses are expressed in this fixed canonical frame.

This representation eliminates global camera motion and allows the model to focus on relative hand dynamics.

\paragraph{Preprocessing.}
Each EgoExo4D clip is converted to the LeRobot~\cite{capuano2025robotlearning, cadenelerobot} sequence format, which synchronizes RGB streams and 3D hand pose annotations. 
For the EgoExo4D dataset, we transformed all 3D coordinates from the world coordinate system to the egocentric coordinate system using extrinsics. During preprocessing, the images were resized to $224\times224$, and the 3D hand positions were normalized using min-max scaling.

\subsection{Baseline Methods}
We compare representative forecasting methods re-evaluated under a unified protocol adapted from the official EgoH4 evaluation, using the same joint definitions, metrics, and prediction horizons:

\begin{itemize}
\item \textbf{Static:} A naive baseline that computes the average 3D pose observed across all samples in the training dataset. This mean pose is then held static and repeated for all future timesteps. It showcases the inherent difficulty and high variance of the forecasting task.
\item \textbf{CVM}~\cite{martinez2017human}: A kinematic baseline that applies constant-velocity extrapolation independently for each joint.
\item \textbf{USST}~\cite{bao2023uncertainty}: A baseline method for predicting future human motion. Since USST does not predict joint-level hand motion, we only compare trajectory metrics.
\item \textbf{EgoH4}~\cite{hatano2025invisible}: A full-protocol forecasting model leveraging body pose and egomotion cues.
\end{itemize}

\subsection{Implementation Details}

\noindent\textbf{Setup.} 
We fine-tune the GR00T-N1.5-3B~\cite{bjorck2025gr00t} model on LeRobot~\cite{capuano2025robotlearning, cadenelerobot} sequences derived from EgoExo4D~\cite{grauman2024ego}.
Each training sample contains synchronized egocentric RGB frames, observed 3D hand poses, and future trajectories.  
Following the EgoH4~\cite{hatano2025invisible} configuration, we use 20 observation frames(2 seconds) to predict 10 future frames(1 second).  
Four RGB frames are uniformly sampled from the observation window as prescribed in EgoH4 and resized to $224{\times}224$ before being processed by the EgoVideo~\cite{pei2024egovideo} encoder.  
Joints with missing annotations are masked during loss computation.

\noindent\textbf{Training.} 
The model is optimized using AdamW with parameters $(\beta_1{=}0.9,\beta_2{=}0.999,\epsilon{=}10^{-8})$.  
We apply a cosine learning-rate schedule with a warm-up ratio of 5\%.  
The final loss function is the weighted sum of the absolute, wrist-relative, and pairwise consistency losses described in Sec.~\ref{sec:training_objectives}.  
We use weighting coefficients $(\lambda_{\text{abs}}, \lambda_{\text{rel}}, \lambda_{\text{pair}}) = (0.6, 0.2, 0.2)$. 

\noindent\textbf{Evaluation Metrics.} 
We evaluate hand motion forecasting using four standard 3D metrics:
Average Displacement Error (ADE), Final Displacement Error (FDE),
Mean Per Joint Position Error (MPJPE), and its final-frame variant (MPJPE-F).
ADE and FDE are computed from the wrist trajectories to measure temporal accuracy,
while MPJPE-based metrics assess spatial consistency across all articulated joints. We report MPJPE-based metrics using wrist-relative MPJPE. All results are averaged over valid frames.
\section{Results}
\subsection{Quantitative Results}

The results are summarized in Tab.~\ref{tab:hand_forecasting}\footnote{We retrained EgoH4 and report the results of the remaining baselines as cited from EgoH4.}. 
Under its original evaluation protocol, EgoH4~\cite{hatano2025invisible} reports ADE = 0.267, FDE = 0.333, MPJPE = 0.116, and MPJPE-F = 0.141. 
Testing EggHand under the same protocol and conditions, it achieves ADE = 0.271, FDE = 0.271, MPJPE = 0.076, and MPJPE-F = 0.077. 
Compared with EgoH4, EggHand achieves comparable ADE (0.271 vs. 0.267) while substantially improving FDE (–18.6\%), MPJPE (–34.5\%), and MPJPE-F (–45.4\%). Although EggHand shows a marginal increase in ADE compared to EgoH4 (0.271 vs. 0.267), it delivers a superior and more balanced error profile by substantially reducing FDE and MPJPE-F.

\begin{table}[t]
    \centering
    \footnotesize
    \setlength{\tabcolsep}{12pt}
    \renewcommand{\arraystretch}{1.15}
    \caption{\textbf{Comparison of methods on hand forecasting tasks}. 
    Left: Hand Trajectory Forecasting, Right: Hand Pose Forecasting. Bold indicates the best results. EggHand achieves a new state-of-the-art in MPJPE and FDE, while maintaining competitive ADE performance relative to EgoH4. Lower is better for all metrics.}
    \label{tab:hand_forecasting}
    {
    \begin{tabular}{lcccc}
        \toprule
        \multirow{2}{*}[-2pt]{\textbf{Method}} 
        & \multicolumn{2}{c}{\textbf{Trajectory}}
        & \multicolumn{2}{c}{\textbf{Pose}} \\
        \cmidrule(lr){2-3} \cmidrule(lr){4-5}
        & \textbf{ADE} & \textbf{FDE}
        & \textbf{MPJPE} & \textbf{MPJPE-F} \\
        \midrule
        Static  & 0.335 & 0.405 & 0.166 & 0.179 \\
        CVM     & 0.346 & 0.467 & 0.166 & 0.183 \\
        USST & 0.562 & 0.581 & - & - \\
        EgoH4   & \textbf{0.267} & 0.333 & 0.116 & 0.141 \\
        \midrule
        \rowcolor{gray!15}
        \textbf{Ours} 
        & 0.271 
        & \textbf{0.271} 
        & \textbf{0.076} 
        & \textbf{0.077} \\
        \bottomrule
    \end{tabular}
    }
\end{table}

\subsection{Ablations}
\begin{table}[t]
    \centering
    \scriptsize
    \setlength{\tabcolsep}{6pt}
    \renewcommand{\arraystretch}{1.15}
    \caption{\textbf{Loss Ablation.} 
    Both $\mathcal{L}_{\text{rel}}$ and $\mathcal{L}_{\text{pair}}$ improve over the $\mathcal{L}_{\text{abs}}$-only baseline, with $\mathcal{L}_{\text{rel}}$ yielding the largest gains. Bold indicates the best results.}
    \label{tab:ablation_loss}
    \label{tab:ablation_loss}

    \label{tab:ablation_loss}
    \resizebox{\linewidth}{!}{
    \begin{tabular}{lcccc}
        \toprule
     \textbf{Method} 
        & \textbf{ADE}
        & \textbf{FDE}
        & \textbf{MPJPE}
        & \textbf{MPJPE-F} \\
        \midrule 
        Ours w. $\mathcal{L}_{\text{abs}}$ & 0.288& 0.289 & 0.077&0.077 \\
        Ours w. $\mathcal{L}_{\text{abs}}+\mathcal{L}_{\text{pair}}$ &0.263 & 0.266 &0.077 &0.077\\
        Ours w. $\mathcal{L}_{\text{abs}}+\mathcal{L}_{\text{rel}}$ & \textbf{0.258} &\textbf{0.259} &\textbf{0.071}  &\textbf{0.071}\\
        \bottomrule
    \end{tabular}
    }
\end{table}

\begin{table}[t]
    \centering
    \scriptsize
    \setlength{\tabcolsep}{6pt}
    \renewcommand{\arraystretch}{1.15}
    \caption{\textbf{Vision Encoder and Input Ablations.} 
Vision offers egocentric grounding and language encodes intent, both of which are essential for accurate forecasting. Bold indicates the best results.}

    \label{tab:ablation_vision}
    \resizebox{\linewidth}{!}{
    \begin{tabular}{lcccc}
        \toprule

     \textbf{Method} 
        & \textbf{ADE}
        & \textbf{FDE}
        & \textbf{MPJPE}
        & \textbf{MPJPE-F} \\
        \midrule 
        Ours w. Noisy Vision & 0.287 & 0.287 & 0.083 & 0.083 \\
        Ours w. Dummy Language & 0.278 & 0.277 & 0.077 & 0.077 \\
        Ours w. Noisy Vision \& Dummy Language & 0.311 & 0.308 & 0.093 & 0.093 \\

        \midrule
        \rowcolor{gray!15}
        \textbf{Ours} 
        & \textbf{0.271} 
        & \textbf{0.271} 
        & \textbf{0.076} 
        & \textbf{0.077} \\
        \bottomrule
    \end{tabular}
    }
\end{table}

\begin{figure*}[t]
  \centering
  \includegraphics[width=1.0\linewidth]{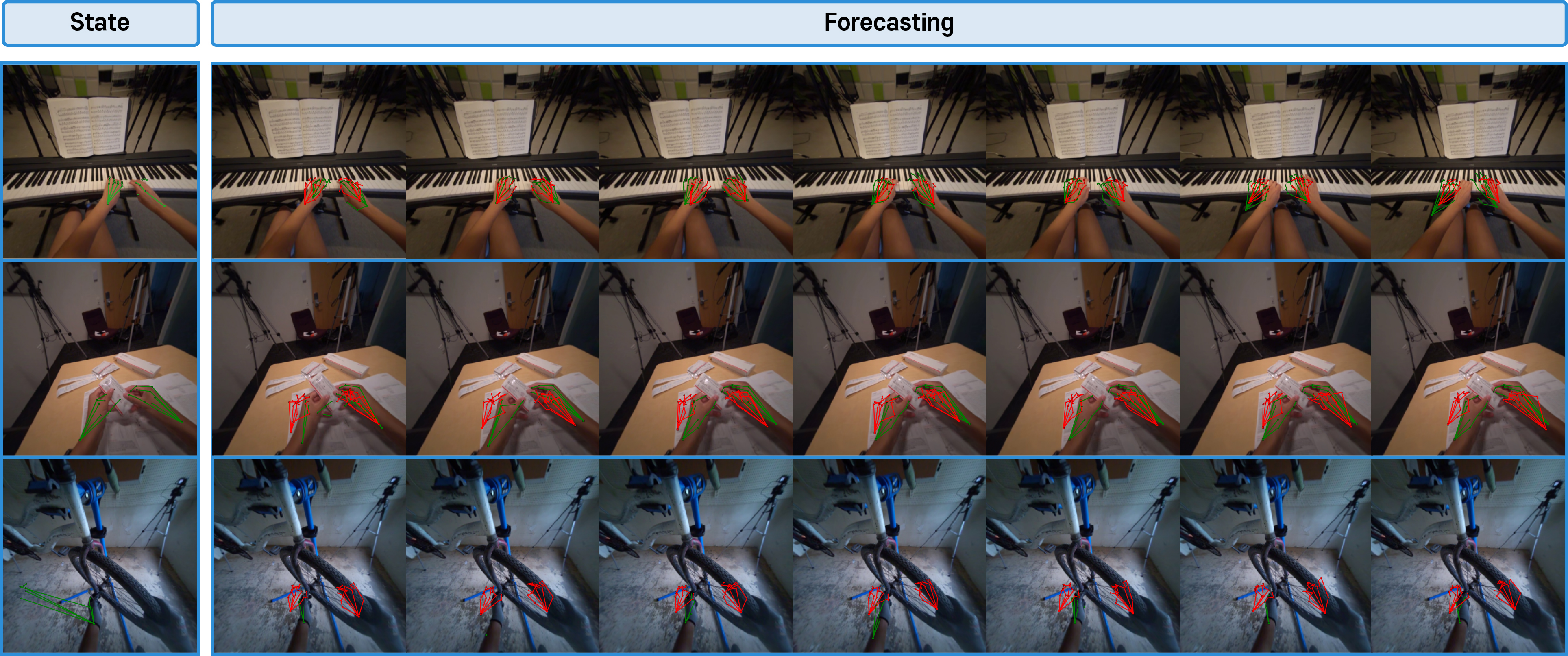}
  \caption{\textbf{Qualitative 2D projections of future 3D hand-pose forecasting on EgoExo4D.}
Green: ground truth; red: our predictions. Left: observation window; right: forecasted future frames.
\textbf{Upper:} piano playing, requiring fine-grained bimanual finger articulation on a structured object.
\textbf{Middle:} COVID-19 rapid antigen test, involving tightly coupled bimanual hand-object interaction.
\textbf{Lower:} bike repair with sparse observations--only one hand is partially visible and future GT is largely missing, yet EggHand still produces plausible trajectories grounded in egocentric scene context.}
  \label{fig:result_qual}
\end{figure*}

\begin{table}[t]
    \centering
    \scriptsize
    \setlength{\tabcolsep}{6pt}
    \renewcommand{\arraystretch}{1.15}
  \caption{\textbf{Egomotion Ratio Analysis.} Comparison of forecasting accuracy across the top egomotion subset (Top 10\%) and the full dataset (All). Bold indicates the best performance. The EgoVideo encoder yields larger gains than InternVL3.5-1B~\cite{chen2024internvl} under high egomotion. Bold indicates the best results.}
    \label{tab:egomotion table}
    \resizebox{\linewidth}{!}{
    \begin{tabular}{lcccc}
        \toprule
     \multirow{2}{*}[-2pt]{\textbf{Method}} 
     & \multicolumn{2}{c}{\textbf{Top 10\%}}
    & \multicolumn{2}{c}{\textbf{All}}
 \\
    
    \cmidrule(lr){2-3}\cmidrule(lr){4-5}
        & \textbf{ADE}
        & \textbf{MPJPE} 
        & \textbf{ADE}
        & \textbf{MPJPE}\\
        \midrule 
        EgoH4 &0.294&0.139&\textbf{0.267}&0.116\\
        Ours w. InternVL3.5-1B &0.303 &0.112 &0.287 &0.082\\
        \midrule 
        \rowcolor{gray!15}
        \textbf{Ours} & \textbf{0.276}& \textbf{0.106} & 0.271&\textbf{0.076}\\
        \bottomrule
    \end{tabular}
    }
\end{table}

\begin{table}[t]
    \centering
    \scriptsize
    \setlength{\tabcolsep}{6pt}
    \renewcommand{\arraystretch}{1.15}
   \caption{\textbf{Effect of VLA Pretraining.} 
We fix the VLM (EgoVideo~\cite{pei2024egovideo}) and vary only the initialization of the GR00T~\cite{bjorck2025gr00t} action head, isolating the impact of VLA-based initialization.}

    \label{tab:ablation_weight}
    \resizebox{\linewidth}{!}{
    \begin{tabular}{lcccc}
        \toprule

     \textbf{Method} 
        & \textbf{ADE}
        & \textbf{FDE}
        & \textbf{MPJPE}
        & \textbf{MPJPE-F} \\
        \midrule 
        Ours(initialized Action Head)&0.321 &0.323 &0.089 &0.089 \\
        \midrule
        \rowcolor{gray!15}
        \textbf{Ours} 
        & \textbf{0.271} 
        & \textbf{0.271} 
        & \textbf{0.076} 
        & \textbf{0.077} \\
        \bottomrule
    \end{tabular}
    }
\end{table}

\subsubsection{Loss Ablations.}
Tab.~\ref{tab:ablation_loss} reports the effect of each auxiliary loss on top of the absolute-coordinate loss $\mathcal{L}_{\text{abs}}$. Adding either the pairwise-distance loss $\mathcal{L}_{\text{pair}}$ or the relative-coordinate loss $\mathcal{L}_{\text{rel}}$ consistently improves over the $\mathcal{L}_{\text{abs}}$-only baseline, with $\mathcal{L}_{\text{rel}}$ yielding the largest gains. This indicates that encoding inter-joint relationships provides complementary supervision to absolute coordinates, helping the model capture the articulated structure of the hand.

\subsubsection{Visual and Language Modality Corruption.}
Tab.~\ref{tab:ablation_vision} shows that replacing the visual input with Gaussian-noise images (``Noisy Vision'') or the language input with randomized dummy text (``Dummy Language'') degrades performance, and corrupting both causes the largest drop. Here, \textit{Dummy Language} denotes randomized task text sampled from the EgoExo4D~\cite{grauman2024ego} task-word vocabulary, and \textit{Noisy Vision} denotes Gaussian-noise images substituted for the original egocentric frames. These results suggest that visual observations and language intent provide complementary information beyond motion priors alone.

\subsubsection{Egomotion Ablation.}
To analyze robustness under strong camera motion, we quantify egomotion as the norm of variation in camera extrinsics across each clip and evaluate on the top 10\% highest-egomotion subset (Tab.~\ref{tab:egomotion table}). On this subset, EggHand consistently outperforms both EgoH4 and our variant equipped with InternVL3.5-1B~\cite{chen2024internvl} across all metrics, indicating that the EgoVideo encoder yields larger gains than a general-purpose VLM encoder under high-egomotion conditions.

\subsubsection{Action Head Pre-trained Weight Ablation.}
Tab.~\ref{tab:ablation_weight} isolates the effect of VLA-pretrained weights on the action head. Fixing the VLM backbone (EgoVideo~\cite{pei2024egovideo}) and varying only the initialization of the GR00T~\cite{bjorck2025gr00t} action head, we find that replacing VLA-pretrained weights with random initialization consistently degrades both trajectory and pose errors, indicating that VLA pretraining provides a strong prior for action prediction.

\begin{figure*}[t]
  \centering
  \includegraphics[width=\linewidth]{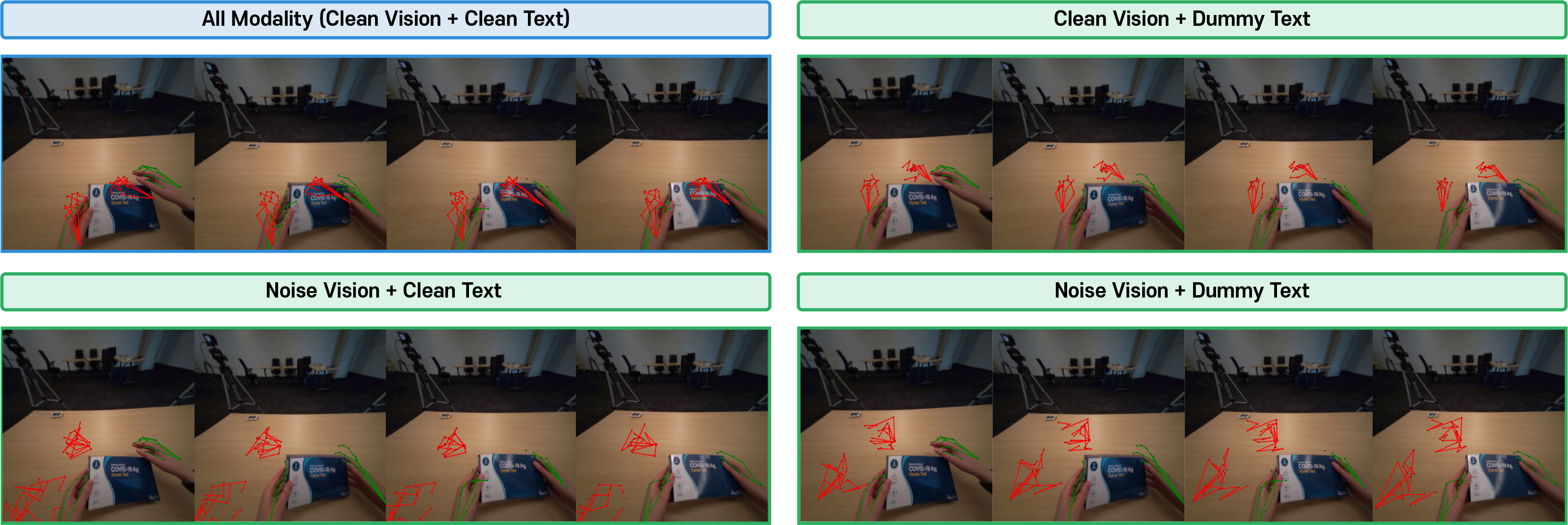}
  \caption{\textbf{Qualitative ablation on multimodal inputs.} EggHand forecasts on a COVID-19 test kit manipulation sequence under four conditions: All Modality, Clean Vision + Dummy Text, Noisy Vision + Clean Text, and Noisy Vision + Dummy Text. Dummy text is randomized from the EgoExo4D task vocabulary; noisy vision replaces frames with Gaussian noise. Green: ground truth; red: predictions.}

  \label{fig:ablation_qual}
\end{figure*}

\subsection{Qualitative Results}
We provide qualitative examples in Fig.~\ref{fig:result_qual} to illustrate EggHand's behavior across tasks.

\textbf{Piano playing (upper).} A two-hand piano performance with both hands visible against the keyboard. EggHand closely follows the fine-grained finger motions across the forecast horizon, anticipating subtle key-press dynamics by grounding its predictions in the egocentric visual context rather than pure motion extrapolation.

\textbf{COVID-19 test kit manipulation (middle).} Handling a rapid antigen test, where each hand's motion is tightly coupled with the grasped object and the other hand. EggHand produces forecasts in which both hands maintain anatomically consistent grasps and move coherently with the object, jointly reasoning about hand pose and hand-object interaction rather than forecasting each hand in isolation.

\textbf{Bike repair with sparse observations (lower).} A bike-repair activity around the handlebar, where only one hand is partially visible and future GT is largely absent---leaving motion-extrapolation baselines with little signal. Nevertheless, EggHand produces anatomically consistent trajectories aligned with the bike, leveraging both egocentric scene cues and language-based task prompts to reconstruct plausible interaction poses even under weak supervision.

\textbf{Qualitative modality ablation (Fig.~\ref{fig:ablation_qual}).} With all modalities intact, EggHand produces stable bimanual grasps aligned with the test kit throughout the forecast horizon. Replacing task text with dummy prompts causes mild drift in hand placement and weakens task-specific finger articulation, indicating that language contributes both task semantics and coarse spatial intent. Replacing egocentric frames with Gaussian noise causes more pronounced drift away from the object and unstable finger configurations, showing that vision is the dominant source of spatial information and geometric consistency. Corrupting both collapses the forecasts into scattered poses spread across the frame. Vision and text thus play complementary roles--vision anchors geometry while text reinforces task intent--and EggHand effectively integrates them.

\section{Conclusion}
We presented EggHand, a multimodal foundation-model framework that couples a VLA action decoder with an egocentric video--text encoder to enable unified reasoning over motion dynamics, semantic intent, and viewpoint shifts for egocentric 3D hand pose forecasting without external tracking. 

On the real-world EgoExo4D benchmark, EggHand achieves state-of-the-art performance in joint-level precision (MPJPE, MPJPE-F) and long-term stability (FDE), while maintaining competitive trajectory performance (ADE). Our egomotion-stratified analysis further shows that on the top 10\% of clips with the largest camera extrinsics variation, EggHand still outperforms both EgoH4 and our InternVL3.5-1B variant, indicating that the EgoVideo encoder provides stable viewpoint-aware grounding where general-purpose VLMs degrade. Ablations clarify the complementary roles of each modality: text drives global placement and task-level articulation, while vision refines local geometry and stabilizes joint evolution. Together, these results suggest that pairing an egocentric-pretrained VLM with an action-structured decoder is a superior paradigm for first-person hand forecasting.

\noindent\textbf{Limitations and future work.}
EggHand consumes past 3D hand poses from an off-the-shelf estimator, and therefore inherits a structural dependency on the upstream module. Nevertheless, our qualitative analysis shows that EggHand remains effective even when future ground-truth annotations are sparse or partially missing due to self-occlusion, producing anatomically consistent and task-aligned predictions by leveraging egocentric visual context and text-based intent cues rather than mere keypoint extrapolation. Promising future directions include end-to-end coupling with the hand pose estimator, extending the prediction horizon, and transferring predicted hand poses to humanoid manipulation policies.
\normalsize
\newpage
\section*{Acknowledgements}

This work was supported by the National Research Foundation of Korea (NRF) grant funded by the Korea government (MSIT) (No. RS-2025-22802992), the Basic Science Research Program through the National Research Foundation of Korea (NRF) funded by the Ministry of Education (No. RS-2025-25420118), the Institute of Information \& Communications Technology Planning \& Evaluation (IITP) grants funded by the Korea government (MSIT) (No. RS-2025-25442149, LG AI STAR Talent Development Program for Leading Large-Scale Generative AI Models in the Physical AI Domain / No. RS-2025-02219277, AI Star Fellowship Support (DGIST)), and the InnoCORE program of the Ministry of Science and ICT (No. N10260003 / No. 26-InnoCORE-01).

{
    \small
    \bibliographystyle{ieeenat_fullname}
    \bibliography{main}
}


\end{document}